\documentclass[lettersize,journal]{IEEEtran}
\usepackage{amsmath,amsfonts}
\usepackage{algorithmic}
\usepackage{algorithm}
\usepackage{array}
\usepackage[caption=false,font=normalsize,labelfont=sf,textfont=sf]{subfig}
\usepackage{textcomp}
\usepackage{stfloats}
\usepackage{url}
\usepackage{verbatim}
\usepackage{graphicx}
\usepackage{hyperref}
\usepackage{cite}
\usepackage{booktabs}   
\usepackage{adjustbox}  
\usepackage{lscape}     
\usepackage{caption}    
\usepackage{xcolor} 
\usepackage{listings}
\lstdefinestyle{cleantext}{
    basicstyle=\ttfamily\small, 
    breaklines=true,
    frame=single,
    framerule=0.5pt,
    backgroundcolor=\color[gray]{0.98},
    showstringspaces=false,
    tabsize=2,
    captionpos=b
}
\begin{document}

\title{Towards Robust Fact-Checking: A Multi-Agent System with Advanced Evidence Retrieval}

\author{
    Tam Trinh\IEEEauthorrefmark{1}\IEEEauthorrefmark{2}, 
    Manh Nguyen\IEEEauthorrefmark{3},
    Truong-Son Hy\IEEEauthorrefmark{2}\thanks{Corresponding author: \href{mailto:thy@uab.edu}{thy@uab.edu}}
    \\
    \IEEEauthorrefmark{1}National Economics University, Vietnam \\
    \IEEEauthorrefmark{2}The University of Alabama at Birmingham, United States \\
    \IEEEauthorrefmark{3}Deakin University, Australia\\
}
\markboth{Journal of \LaTeX\ Class Files,~Vol.~14, No.~8, August~2021}%
{Shell \MakeLowercase{\textit{et al.}}: A Sample Article Using IEEEtran.cls for IEEE Journals}


\maketitle

\begin{abstract}
The rapid spread of misinformation in the digital era poses significant challenges to public discourse, necessitating robust and scalable fact-checking solutions. Traditional human-led fact-checking methods, while credible, struggle with the volume and velocity of online content, prompting the integration of automated systems powered by Large Language Models (LLMs). However, existing automated approaches often face limitations, such as handling complex claims, ensuring source credibility, and maintaining transparency. This paper proposes a novel multi-agent system for automated fact-checking that enhances accuracy, efficiency, and explainability. The system comprises four specialized agents: an Input Ingestion Agent for claim decomposition, a Query Generation Agent for formulating targeted subqueries, an Evidence Retrieval Agent for sourcing credible evidence, and a Verdict Prediction Agent for synthesizing veracity judgments with human-interpretable explanations. Evaluated on benchmark datasets (FEVEROUS, HOVER, SciFact), the proposed system achieves a 12.3\% improvement in Macro F1-score over baseline methods. The system effectively decomposes complex claims, retrieves reliable evidence from trusted sources, and generates transparent explanations for verification decisions. Our approach contributes to the growing field of automated fact-checking by providing a more accurate, efficient, and transparent verification methodology that aligns with human fact-checking practices while maintaining scalability for real-world applications. Our source code is available at \url{https://github.com/HySonLab/FactAgent}.
\end{abstract}


\section{Introduction}
The rapid proliferation of misinformation on digital platforms poses a significant threat to the integrity of information and public discourse. As false or misleading content spreads at unprecedented speed, the demand for effective and scalable fact-checking solutions has become increasingly urgent \cite{10.1145/3442188.3445922}. Although traditional fact-checking performed by human experts and specialized organizations has played a vital role, it is still insufficient to keep up with the sheer volume and velocity of misinformation in the modern media landscape \cite{Graves2018UnderstandingTP}.

Misinformation, defined as false information presented as factual regardless of intent, has far-reaching consequences in the political, health, and economic domains. In political settings, it can distort democratic processes and exacerbate social polarization \cite{doi:10.1126/science.aap9559}. During public health crises such as the COVID-19 pandemic, misinformation has contributed to vaccine hesitancy and the spread of harmful behaviors \cite{doi:10.1177/0268580920914755}. In economic contexts, unverified claims have been linked to market volatility and misguided consumer decisions \cite{10.1016/j.ipm.2019.03.004}.

To address these issues at scale, recent efforts have turned to automated fact-checking systems, bolstered by advances in LLMs \cite{zhao2023survey}. These models exhibit impressive natural language understanding and generation capabilities, enabling them to process and evaluate claims on a previously unattainable scale. However, deploying LLMs for fact-checking also introduces new challenges. LLMs are prone to hallucinations, generating plausible but incorrect information and often operate as black boxes, making it difficult to interpret or verify their output \cite{10.1145/3571730}. Furthermore, LLMs trained in static corpora may lack access to up-to-date or domain-specific knowledge required for accurate verification \cite{Feng2023KnowledgeST}.

To mitigate these limitations, researchers have explored hybrid strategies that integrate LLMs with retrieval mechanisms and structured reasoning pipelines. Notable approaches include retrieval-augmented generation \cite{lewis2020retrieval}, chain-of-thought prompting \cite{wei2022chain}, and tool-based reasoning frameworks \cite{chern2023factool}. Despite these advancements, current systems often fall short when handling complex, multi-faceted claims, filtering non-verifiable statements, and ensuring the credibility and completeness of retrieved evidence. Many models also rely on shallow web snippets rather than full-document analysis, limiting the depth and reliability of their verifications \cite{10.1145/3137597.3137600}.

\begin{figure}[htbp]
    \centering
    \includegraphics[width=\linewidth]{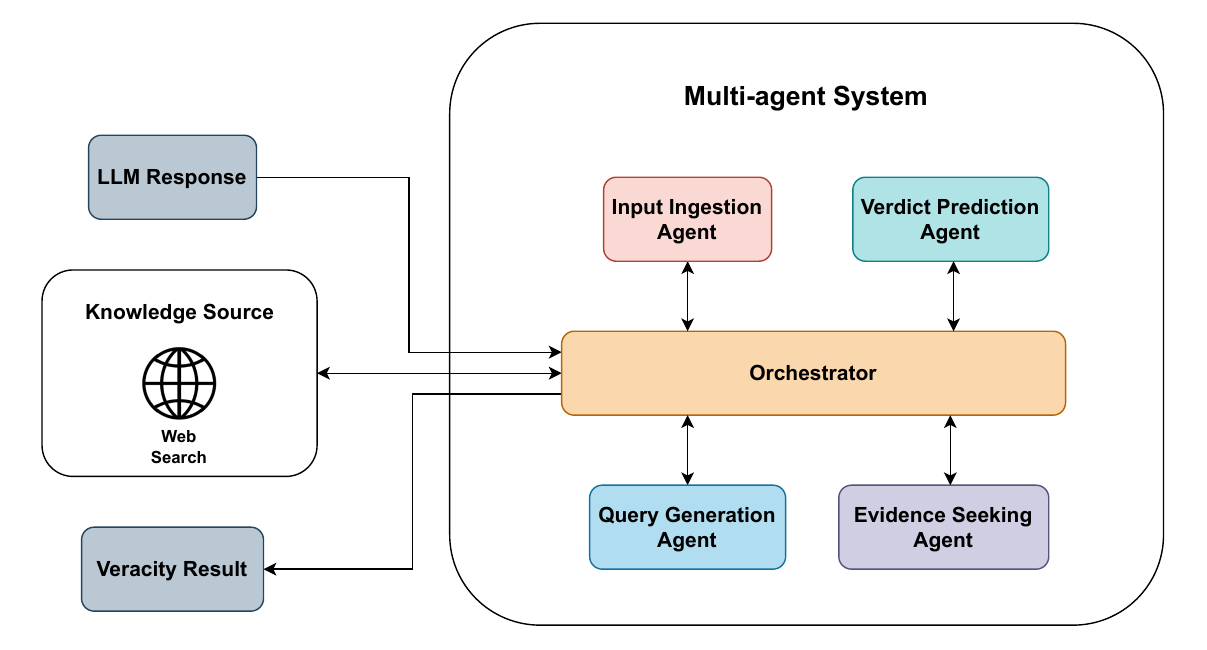}
    \caption{Overview of the multi-agent system. The system comprises an Orchestrator that manages communication and workflow between specialized agents, including Input Ingestion Agent, Query Generation Agent, Evidence Seeking Agent, and Verdict Prediction Agent. The Orchestrator interacts with a knowledge source and integrates their responses to produce a final Veracity Result.}
    \label{fig:my-overview}
\end{figure}

In this paper, we propose a novel multi-agent framework for automated fact-checking that integrates program-guided reasoning, high-quality evidence retrieval, and interpretable verdict generation. Our system is composed of four specialized agents (Figure \ref{fig:my-overview}):

\begin{itemize}
    \item \textbf{Input Ingestion Agent:} Decomposes complex claims into sub-claims and identifies their verifiability.
    \item \textbf{Query Generation Agent:} Constructs targeted search queries to retrieve relevant information.
    \item \textbf{Evidence Retrieval Agent:} Sources of credible information from trusted knowledge bases and documents.
    \item \textbf{Verdict Prediction Agent:} Synthesizes retrieved evidence to assess the validity of the claim and generate explanations.
\end{itemize}

The key contributions of this work are as follows.

\begin{enumerate}
    \item We introduce a structured multi-agent architecture that improves fact-checking precision through decomposable reasoning and modular information processing.
    \item We implement a robust evidence retrieval strategy that emphasizes source credibility and leverages full-document content rather than relying solely on search engine snippets.
    \item We conducted extensive experiments on three benchmark datasets including FEVEROUS \cite{Aly21Feverous}, HOVER \cite{jiang-etal-2020-hover}, and SciFact \cite{wadden-etal-2020-fact}, demonstrating that our system achieves superior performance compared to existing baselines by 12.3\% relative improvement.
\end{enumerate}

\section{Related Works}

\subsection{Traditional Fact-Checking Approaches}

Fact-checking has its roots in traditional journalism, where professional fact-checkers manually verified claims using rigorous investigation and source analysis \cite{Graves2018UnderstandingTP}. Organizations like PolitiFact \cite{politifact}, Snopes \cite{snopes}, and FactCheck.org \cite{factcheck} pioneered standardized methodologies, particularly for political discourse \cite{amazeen2015revisiting}. However, these manual approaches face serious scalability issues in the digital age, where content is produced and distributed at unprecedented speed.

The emergence of semi-automated fact-checking systems marked the beginning of computational support for human experts. Vlachos and Riedel~\cite{Vlachos2014FactCT} conceptualized fact-checking as a classification task and laid the groundwork for early benchmarks. Subsequently, natural language processing (NLP) techniques were incorporated to analyze claims and retrieve relevant evidence. In particular, Hassan et al.~\cite{hassan2017toward} proposed ClaimBuster, a pipeline that identifies checkworthy claims and assists in verification, demonstrating the viability of human-AI collaboration in this domain.

\subsection{Automated Fact-Checking Systems}
Automated fact-checking systems can generally be divided into content-based and context-based approaches~\cite{10.1145/3442188.3445922}. Content-based methods analyze linguistic and semantic features within text to detect misinformation, while context-based methods focus on how information spreads, which examine propagation patterns, user interactions, and temporal signals.

Advancements in neural architectures have substantially improved system performance. Popat et al.~\cite{popat2018declare} proposed a deep learning model that evaluates the credibility of claims by analyzing the position and reliability of evidence. Augenstein et al.~\cite{augenstein-etal-2019-multifc} introduced MultiFC, a benchmark dataset that emphasizes the importance of verifying claims against multiple and diverse sources. Transformer-based architectures further pushed the boundaries, with models like GEAR~\cite{zhou-etal-2019-gear} leveraging graph-based aggregation to synthesize multi-evidence reasoning.

\subsection{LLMs for Fact-Checking}

LLMs have recently become a powerful tool for fact-checking due to their capabilities in language understanding and generation~\cite{zhao2023survey}. Initial LLM-based methods used zero-shot and few-shot prompting to evaluate claims without requiring extensive fine-tuning~\cite{Stiff2021DetectingCD}. Lee et al.~\cite{lee-etal-2021-towards} showed that LLMs could generate natural language explanations, enhancing interpretability.

However, LLMs suffer from several critical limitations. A major issue is hallucination, which generates plausible but incorrect information~\cite{10.1145/3571730}. Studies such as~\cite{DU2023QuantifyingAA} have documented how LLMs confidently produce inaccurate results, undermining their trustworthiness. Moreover, static training data results in temporal obsolescence, hindering the verification of claims about recent or dynamic events.

To mitigate these issues, researchers have proposed several enhancements. Retrieval-augmented generation (RAG)~\cite{lewis2020retrieval} grounds LLM outputs using external evidence. Chain-of-thought prompting~\cite{wei2022chain} improves the transparency of reasoning by encouraging step-by-step explanations. Tool-augmented approaches such as FacTool~\cite{chern2023factool} equip LLMs with specialized modules, improving domain robustness and factuality.

\subsection{Multi-Agent Systems for Complex Reasoning}

Multi-agent systems (MAS) have emerged as a promising architecture to enhance LLM reasoning through task decomposition and modular specialization~\cite{xi2023rise}. In fact-checking, MAS approaches distribute responsibilities such as claim parsing, query generation, evidence retrieval, and verdict synthesis across individual agents, improving scalability and interpretability.

Li et al.~\cite{li2023can} demonstrated that debate-style MAS can enhance decision faithfulness and reduce hallucinations. Similarly, Zhang et al.~\cite{zhang2023web} introduced retrieval agents that autonomously browse the web to collect real-time evidence, increasing adaptability to evolving misinformation. Self-verification has also been explored; Wang et al.~\cite{wang2023self} proposed \textit{Self-Checker}, which empowers LLMs to critique their own outputs using structured reasoning and external tools, thus reducing reliance on static knowledge.

Despite these advances, challenges persist. Effective MAS require robust inter-agent communication, consistency in distributed reasoning, and efficient task allocation. Our proposed system builds on these insights, introducing a structured multi-agent framework that prioritizes decomposable reasoning, credible source retrieval, and transparent explanation generation.

\section{Research Methodology}
\begin{figure*}[htbp]
    \centering
    \includegraphics[width=\textwidth]{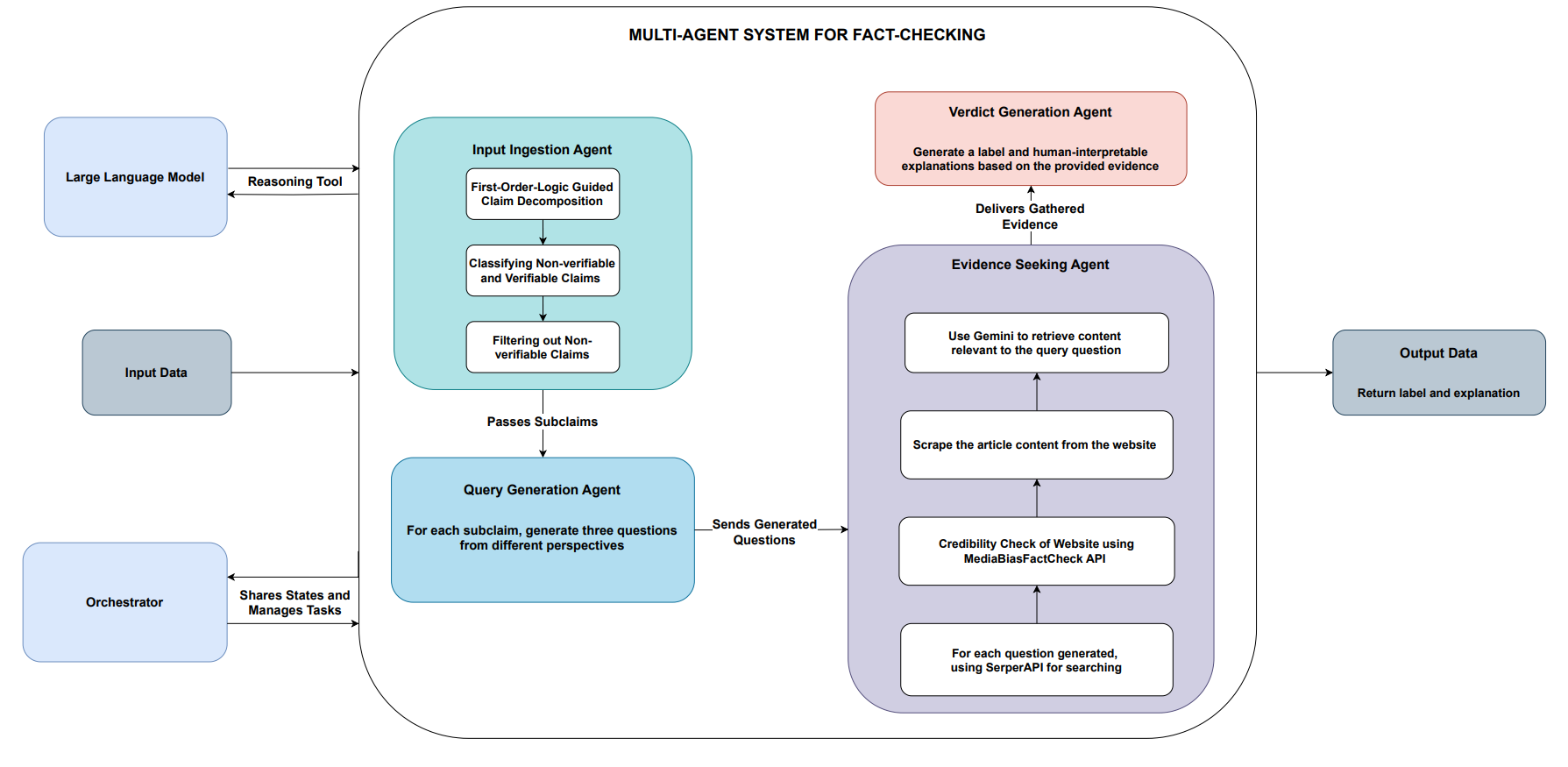}
    \caption{Overview of the proposed architecture. The system takes input data and decomposes claims using Input Ingestion Agent. Verifiable subclaims are passed to Query Generation Agent, which creates questions from different perspectives. Evidence Seeking Agent retrieves and verifies evidence from web sources. Finally, Verdict Generation Agent produces a fact-check label and explanation based on the gathered evidence.}
    \label{fig:fact-checking-pipeline}
\end{figure*}

The proposed multi-agent system for automated fact-checking comprises four specialized agents, including Input Ingestion Agent, Query Generation Agent, Evidence Seeking Agent, and Verdict Prediction Agent, each responsible for a distinct stage in the claim verification pipeline. Each agent is powered by LLM, leveraging their advanced reasoning capabilities through carefully engineered prompts that guide their behavior toward specific verification tasks. As illustrated in Figure \ref{fig:fact-checking-pipeline}, these agents based on LLM operate sequentially to transform raw input claims into structured verified outputs accompanied by human-interpretable explanations.
\subsection{Input Ingestion Agent}
The Input Ingestion Agent serves as the entry point for the fact-checking pipeline, processing raw claims, and preparing them for downstream verification. This agent performs two critical functions: decomposing complex claims into atomic components using First-Order Logic (FOL) and filtering out non-verifiable statements.
\subsubsection{Decomposing into FOL Subclaims}
Complex claims often contain multiple propositions, implicit assumptions, or conditional dependencies that can hinder effective verification. To address this challenge, the Input Ingestion Agent transforms multi-faceted claims into simplified sub-claims using techniques inspired by First-Order Logic representation \cite{enderton2001mathematical}.

According to Enderton \cite{enderton2001mathematical}, FOL extends propositional logic by introducing predicates that describe properties of objects or relations between them. A well-formed FOL formula consists of atomic predicates applied to terms (constants or variables), formally expressed as $P(t_1, t_2, ..., t_n)$ where $P$ is a predicate of arity $n$ and $(t_1, t_2, ..., t_n)$ are terms.
For example, the claim \textit{"Sumo wrestler Toyozakura Toshiaki committed match-fixing, ending his career in 2011 that started in 1989"} is decomposed into:

\begin{itemize}
    \item \texttt{Occupation(Toyozakura Toshiaki, "sumo wrestler")}
    \item \texttt{Commit(Toyozakura Toshiaki, "match-fixing")}
    \item \texttt{Ending(Toyozakura Toshiaki, "his career in 2011")}
    \item \texttt{Starting(Toyozakura Toshiaki, "his career in 1989")}
\end{itemize}

Each predicate is accompanied by a natural language verification goal (e.g., \textit{"Verify that Toyozakura Toshiaki is or was a sumo wrestler"}). At this stage, for a given input claim $C$, Input Ingestion Agent initially generates a collection of predicates, denoted as $\mathcal{P} = [p_1, \dots, p_n]$. These predicates correspond to the constituent sub-claims $C = [c_1, \dots, c_n]$. Each predicate $p_i \in \mathcal{P}$ is an FOL expression designed to guide LLMs in formulating a question-and-answer pair that encapsulates the meaning of the sub-claim $c_i$. The general claim $C$ can be conceptualized as a conjunction of these predicates, that is, $C = p_1 \wedge p_2 \wedge \dots \wedge p_n$. To establish that claim $C$ is SUPPORTED, every predicate must be evaluated as True. Conversely, if even one predicate is found to be False, the claim is categorized as NON\_SUPPORTED.

The decomposition facilitates granular reasoning, its effectiveness has been proven by recent research in automated fact verification by Wang and Shu ~\cite{wang-shu-2023-explainable}. Wang and Shu \cite{wang-shu-2023-explainable} found that FOL-guided decomposition enables more precise evidence retrieval and reduces error propagation in multi-hop reasoning tasks.

\subsubsection{Filtering out Non-verifiable Subclaims}
Following decomposition, the agent classifies each subclaim as verifiable or non-verifiable based on established criteria from Micallef et al. \cite{micallef2022claim} and Konstantinovskiy et al. \cite{konstantinovskiy2021claim}. Verifiable claims are factual statements that can be independently checked for accuracy using objective evidence, while non-verifiable claims express subjective opinions or lack specificity for objective verification.

The classification process considers several key characteristics: \begin{itemize} 
    \item \textbf{Verifiable claims} involve checkable facts or statistics, can be verified using trusted sources, and are objective in nature. 
    \item \textbf{Non-verifiable claims} are often subjective, emotional, or anecdotal; cannot be objectively proven or disproven; and depend on personal belief or preference. 
\end{itemize}

After classification, non-verifiable claims are filtered out from further processing. This filtering ensures that only claims that can be validated with factual evidence proceed to subsequent stages, preventing unnecessary computations and focusing the system's resources on statements that can be objectively evaluated.
\subsection{Query Generation Agent}

The Query Generation Agent transforms atomic subclaims into effective search queries designed to retrieve relevant evidence. This agent leverages the reasoning capabilities of LLMs to generate diverse, well-formulated questions that maximize the likelihood of retrieving accurate information from search engines. The development of this module draws inspiration from previous work~\cite{press-etal-2023-measuring, wang-shu-2023-explainable}, which explores the generation of queries to support evidence-based reasoning. However, the cited work does not provide detailed methodologies for crafting specific questions, leaving a gap in practical implementation. To address this, Query Generation Agent incorporates principles from Search Engine Optimization (SEO) to guide the formulation of questions.

For each atomic subclaim produced by the Input Ingestion Agent, the Query Generation Agent creates $k$ distinct search queries. These queries are deliberately formulated to approach the verification task from multiple angles, ensuring comprehensive evidence collection while mitigating potential biases in search results. The agent incorporates principles from Search Engine Optimization (SEO) research (\cite{baye2016search, 7977137}), including: 
\begin{itemize} 
    \item Use of specific keywords and entities from the claim 
    \item Incorporation of synonyms and alternative phrasings 
    \item Balanced specificity to maximize relevance without over-constraining results 
    \item Question-based formulations that align with natural search patterns 
\end{itemize}

This multi-perspective generation strategy ensures that the downstream Evidence Seeking Agent is equipped with varied entry points for retrieving supporting or refuting material from the website. 
\subsection{Evidence Seeking Agent}

The Evidence Seeking Agent is responsible for retrieving, validating, and extracting relevant information from web sources. This agent incorporates tools for programmatic search, credibility assessment, and targeted content extraction to ensure that the prediction of the following verdict is based on comprehensive and reliable evidence. To the best of our knowledge, our work is the first to propose an Evidence Seeking Agent that integrates programmatic web search with systematic verification of the credibility of the media site and comprehensive extraction of evidence from retrieved content.
\subsubsection{Internet Search}
The initial stage of evidence gathering involves programmatic access to search engine results to identify sources relevant to the queries generated. For this purpose, we use SerperAPI \cite{serper2024}, a high-performance and cost-effective Google Search API. 

Our implementation configures SerperAPI with parameters optimized for fact-checking purposes: 
\begin{itemize} 
\item \textbf{Result limitation} (\texttt{num=10}): We retrieve the top 10 search results for each query to optimize cost efficiency. 
\item \textbf{Regional targeting} (\texttt{gl=us}): Searches are configured for the US region to ensure consistency in the result patterns. \item \textbf{Temporal boundaries} (\texttt{tbs}): To prevent temporal leakage when evaluating against benchmark datasets, we implement dataset-specific end-date restrictions aligned with publication dates:      
\begin{itemize} 
    \item FEVEROUS \cite{Aly21Feverous}: October 12, 2021 
    \item HoVer \cite{jiang-etal-2020-hover}: November 16, 2020 
    \item SciFact \cite{wadden-etal-2020-fact}: October 3, 2020 \end{itemize} 
\end{itemize}

This temporal bounding is particularly important for maintaining methodological integrity in fact-checking evaluation, as it prevents the system from accessing information that would not have been available at the time each benchmark dataset was created.
\subsubsection{Credibility Check in Media Sites}
Distinguishing between reliable and unreliable information sources is critical in the fact-checking process. 
For source credibility assessment, we integrate the Media Bias/Fact Check (MBFC) API \cite{mediabiasfactcheck}, which provides programmatic access to professional human ratings of publisher credibility. The system filters potential evidence sources based on specific thresholds for the quality of factual reporting and political bias (Table \ref{tab:publisher_categorization}). 
\begin{table}[ht]
\small 
\setlength{\tabcolsep}{4pt} 
\renewcommand{\arraystretch}{1.1} 
\centering
\begin{tabular}{|p{1.5cm}|p{2.8cm}|p{2.8cm}|}
\hline
\textbf{Category} & \textbf{Before Filtering} & \textbf{After Filtering} \\
\hline
\textbf{Factuality} & 
\begin{itemize}
  \item ``very high''
  \item ``high''
  \item ``mostly factual''
  \item ``mixed''
  \item ``low''
  \item ``very low''
\end{itemize} & 
\begin{itemize}
  \item ``very high''
  \item ``high''
  \item ``mostly factual''
\end{itemize} \\
\hline
\textbf{Political Bias} & 
\begin{itemize}
  \item ``least biased''
  \item ``left-center''
  \item ``right-center''
  \item ``left''
  \item ``right''
  \item ``extremely left''
  \item ``extremely right''
  \item ``pro-science''
  \item ``questionable''
  \item ``satire''
  \item ``conspiracy-pseudoscience''
\end{itemize} & 
\begin{itemize}
  \item ``least biased''
  \item ``left-center''
  \item ``right-center''
  \item ``pro-science''
\end{itemize} \\
\hline
\end{tabular}
\caption{Publisher categorization before and after filtering in MBFC.}
\label{tab:publisher_categorization}
\end{table}

For domains not listed in the MBFC database, we implement a fallback credibility assessment mechanism that considers the following: 
\begin{itemize} 
    \item \textbf{Domain suffix analysis}: Academic domains (.edu), government domains (.gov), and established organizational domains (.org) receive higher credibility scores \cite{METZGER2013210}. 
    \item \textbf{Publication history}: Domains with established publication histories are considered more reliable than recently created websites \cite{dong2015knowledge}. 
    \item \textbf{Citation patterns}: For scientific claims, sources appearing in citation databases such as Google Scholar or PubMed are prioritized. 
\end{itemize}
\subsubsection{Evidence Retrieval}
This content extraction process marks a substantial improvement over previous methodologies that rely solely on search result snippets (\cite{Zhou2024CorrectingMO, zhang-gao-2023-towards, wang-shu-2023-explainable}), which often lack the necessary context for thorough fact verification.

Our method adopts a robust two-stage content extraction strategy:

\begin{enumerate}
    \item \textbf{Comprehensive full-text retrieval:} We utilize Selenium~\cite{selenium2023} for browser automation in conjunction with BeautifulSoup~\cite{richardson2007beautiful} for HTML parsing. This integration supports full JavaScript rendering, accurate element targeting via CSS selectors and XPath expressions, and seamless navigation across dynamic web pages.

    \item \textbf{Contextually relevant passage identification:} Using Google's Gemini 1.5 Flash model~\cite{team2024gemini}, which features an expansive context window of approximately one million tokens, the system is capable of processing and analyzing content from multiple articles simultaneously.
\end{enumerate}

For each query, the system selects the top-ranked credible source and extracts the full textual content. A custom prompt then guides the LLM to isolate only the most pertinent passages, ensuring relevance to the original subclaim.

The extracted passages are stored in a structured evidence repository, along with associated metadata, including the source URL, credibility rating, and extraction timestamp. This repository serves as the primary input for the subsequent \textit{Verdict Prediction Agent}.

\subsection{Verdict Prediction Agent}
The Verdict Prediction Agent serves as the final component in our multi-agent fact-checking pipeline. It is responsible for synthesizing the evidence collected by the Evidence Seeking Agent and determining the veracity of each decomposed subclaim. This agent evaluates the credibility, consistency and relevance of the evidence collected to deliver a final verdict either \texttt{supported} or \texttt{not\_supported}, along with a human-interpretable explanation.

The decision-making process follows a structured, multi-step methodology:

\begin{enumerate}
    \item \textbf{Evidence analysis:} The agent examines all retrieved evidence related to each subclaim and assesses the degree of agreement between multiple sources. A higher level of consistency across diverse and credible sources increases confidence in the resulting verdict.

    \item \textbf{Structured voting mechanism:} A weighted voting system is applied, where multiple strong pieces of supporting evidence lead to a \texttt{supported} verdict. In contrast, the presence of contradictory or insufficient reliable evidence results in a \texttt{not\_supported} verdict.

    \item \textbf{Explanation generation:} For each decision, the agent generates a detailed explanation that references specific evidence and describes the reasoning process. This ensures both transparency and interpretability of the output.
\end{enumerate}


\section{Experiments}
\subsection{Datasets}
To evaluate our multi-agent fact-checking system, we conducted experiments on three of the most widely recognized and popular benchmark datasets in fact verification. We employed stratified sampling to select 100 examples from each dataset, ensuring balanced label distribution while managing computational costs. Our system leverages a zero-shot, training-free approach, which mitigates this constraint by enabling robust generalization without the need for extensive fine-tuning or large-scale training data.
\begin{itemize}
    \item \textbf{HoVER} \cite{jiang-etal-2020-hover}: Designed to challenge fact verification systems with multi-hop reasoning, HoVER requires connecting multiple pieces of evidence across Wikipedia articles to verify claims. We stratified the samples of the validation set into two-, three- and four-hop claims to rigorously evaluate our system’s ability to handle increasing complexity of reasoning.
    \item \textbf{FEVEROUS} \cite{Aly21Feverous}: This benchmark tests complex claim verification using both structured (tables) and unstructured (text) data from Wikipedia, covering numerical, multi-hop, and mixed-format reasoning challenges. We selected validation set claims to assess the versatility of our framework across these diverse reasoning types critical to real-world fact checking.
    \item \textbf{SciFact-Open} \cite{wadden-etal-2020-fact}: Focused on open-domain scientific claim verification, this dataset demands retrieving and evaluating evidence from the specialized literature without predefined veracity labels for all claims. We curated validation set claims with complete supporting or contradictory evidence, testing our system’s precision in a domain requiring deep knowledge.
\end{itemize}

\subsection{Baseline Approaches}

We compare our multi-agent fact-checking framework against four representative baseline methods, each reflecting a distinct paradigm in automated claim verification.

\textbf{Direct} approach employs a simple closed-book fact-checking strategy. An LLM receives the input claim and directly produces a veracity label (\texttt{true}, \texttt{false}, or \texttt{partially true}) along with an explanatory rationale in a single inference pass. This method relies entirely on the model's internal parametric knowledge without any external evidence retrieval. Although straightforward to implement, this baseline is limited by potential knowledge staleness, hallucination risks, and the lack of transparency in its reasoning process.

\textbf{Chain-of-Thought (CoT)} method, introduced by Wei et al. \cite{10.5555/3600270.3602070}, enhances the model reasoning through structured prompting that encourages sequential step-by-step inference. For fact verification tasks, CoT is applied as a two-stage pipeline: (1) decomposition of the claim into a set of verification sub-questions, and (2) iterative reasoning over each sub-question to reach a final verdict. Although this improves interpretability via intermediate reasoning steps, the approach is vulnerable to cascading errors and depends critically on the quality of the initial decomposition.

\textbf{Self-Ask with Search Engine (SA+SE)}, based on the Self-Ask framework \cite{press-etal-2023-measuring}, introduces a formalized decomposition strategy augmented by real-time knowledge retrieval. When the LLM generates a follow-up sub-question during reasoning, the system intercepts it and queries an external search engine. The retrieved content is then provided as an intermediate answer to inform the subsequent reasoning. This hybrid strategy combines parametric reasoning with external factual grounding, though it is limited by the model's ability to formulate effective search queries and its lack of built-in mechanisms for evaluating source credibility.

\textbf{FOLK} framework \cite{wang-shu-2023-explainable} integrates symbolic reasoning with external evidence grounding. It operates in three key stages: (1) decomposition of claims into FOL predicates, representing atomic factual assertions; (2) external knowledge grounding via search queries associated with each predicate; and (3) prediction of predicate-level veracity based on the retrieved knowledge.

Although FOLK offers a high degree of interpretability and reduces hallucination risk by grounding each sub-claim in external data, it is hindered by several shortcomings. These include redundant predicate generation, reliance on shallow evidence from search snippets rather than full-document content, and the absence of robust source credibility assessment, all of which can undermine performance in verifying complex or controversial claims.

\subsection{Experimental Settings}
We evaluated our multi-agent system using open source models (Llama-3.2-1B\footnote{\url{https://ollama.com/library/llama3.2:1b}} and Qwen-2.5-3B\footnote{\url{https://ollama.com/library/qwen2.5:3b}}) deployed locally via Ollama\footnote{\url{https://ollama.com/}} and commercial models (GPT-4o-mini\footnote{\url{https://openai.com/index/gpt-4o-mini-advancing-cost-efficient-intelligence/}}) accessed through API services. LangGraph\footnote{\url{https://www.langchain.com/langgraph}} served as the orchestration framework, coordinating the workflow between agents and managing state transitions throughout the verification pipeline. For evidence retrieval, we integrate Gemini-1.5-flash
\footnote{\url{https://ai.google.dev/gemini-api/docs/models}}
to process the complete content of the document. We systematically varied the number of generated questions per subclaim (1-5) to optimize the balance between verification thoroughness and computational efficiency. All experiments were carried out on stratified samples from HoVER, FEVEROUS, and SciFact-Open datasets with Macro F1 score as metrics in all configurations.

\subsection{Main Results}
Table~\ref{tab:results} presents a comprehensive performance comparison between our proposed multi-agent system and four established baseline approaches: Direct, CoT, SA+SE, and FOLK. All systems were implemented using GPT-4o-mini as the foundation model, with our multi-agent system configured to generate 3 queries per subclaim.
\begin{table*}[ht]
\centering
\begin{adjustbox}{width=0.9\textwidth}
\begin{tabular}{l|ccc|ccc|c}
\toprule
\textbf{Method} & \multicolumn{3}{c|}{\textbf{HoVER}} & \multicolumn{3}{c|}{\textbf{FEVEROUS}} & \textbf{SciFact-Open} \\
& \textbf{2-Hop} & \textbf{3-Hop} & \textbf{4-Hop} & \textbf{Numerical} & \textbf{Multi-hop} & \textbf{Text+Table} &  \\
\midrule
Direct & 0.521 & 0.449 & 0.419 & 0.465 & 0.502 & 0.591 & 0.497 \\
CoT & 0.54 & 0.466 & 0.44 & 0.496 & 0.609 & 0.618 & 0.634 \\
SA+SE & 0.542 & 0.489 & 0.46 & 0.553 & 0.612 & 0.542 & 0.609 \\
FOLK & 0.595 & 0.501 & 0.466 & 0.487 & \textbf{0.630} & 0.649 & 0.737 \\
\textbf{MAS (Ours)} & \textbf{0.600 (+0.005)} & \textbf{0.617 (+0.116)} & \textbf{0.507 (+0.041)} & \textbf{0.548 (+0.061)} & 0.601 & \textbf{0.681 (+0.031)} & \textbf{0.770 (+0.033)} \\
\bottomrule
\end{tabular}
\end{adjustbox}
\caption{Macro F1-score of baseline methods and our Multi-Agent System (MAS) across the HoVER, FEVEROUS, and SciFact-Open datasets using GPT-4o-mini. The number of questions generated by the Query Generation Agent is 3. \textbf{Bold} indicates the best result in each column. MAS outperforms existing baselines in 6 out of 7 evaluation settings.}
\label{tab:results}
\end{table*}
Our proposed MAS achieves superior performance in six of seven evaluation tasks, as shown in Table~\ref{tab:results}, demonstrating its robustness against diverse reasoning challenges in claim verification. Two primary observations highlight its effectiveness.

First, MAS excels in complex multi-hop reasoning, particularly on the HoVER dataset. For 3-hop and 4-hop claims, MAS achieves F1-scores of 0.617 and 0.507, surpassing the best baseline (FOLK) by 23.15\% and 8.80\%, respectively. This advancement stems from MAS’s structured pipeline which comprises four specialized agents including Input Ingestion, Query Generation, Evidence Retrieval, and Verdict Prediction. By decomposing complex claims into verifiable subclaims and generating targeted queries, MAS improves evidence retrieval and synthesis accuracy. Unlike baselines such as Direct and CoT which falter in deeper reasoning tasks, MAS’s FOL-guided approach ensures robust handling of intricate dependencies, improving reasoning fidelity.

Second, MAS demonstrates exceptional performance in diverse evidence contexts, notably in FEVEROUS’s Text+Table and SciFact-Open tasks, with F1-scores of 0.681 and 0.770, respectively. These results reflect a 4.93\% relative improvement over FOLK in Text+Table and a 4.48\% relative improvement in SciFact-Open, driven by MAS’s ability to prioritize credible sources in real time. Unlike less structured approaches like SA+SE and Direct, MAS integrates FOL to ensure consistent verdict synthesis across heterogeneous evidence types, minimizing errors in numerical and multi-modal claims.
\subsection{Impact of Number of Generated Questions}
\begin{figure*}[ht]
  \centering
  \makebox[\textwidth][c]{%
    \begin{minipage}{1.1\textwidth}
      \centering
      \includegraphics[width=.33\linewidth]{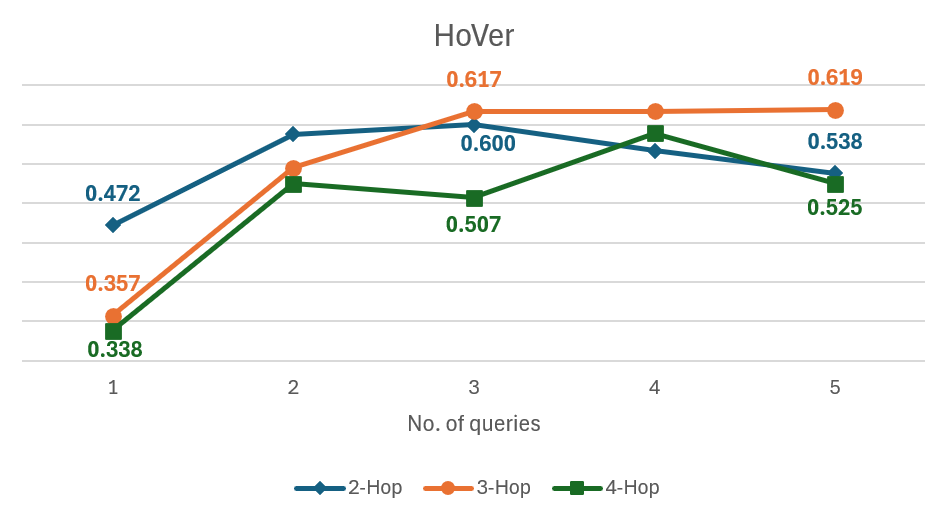}%
      \hfill
      \includegraphics[width=.33\linewidth]{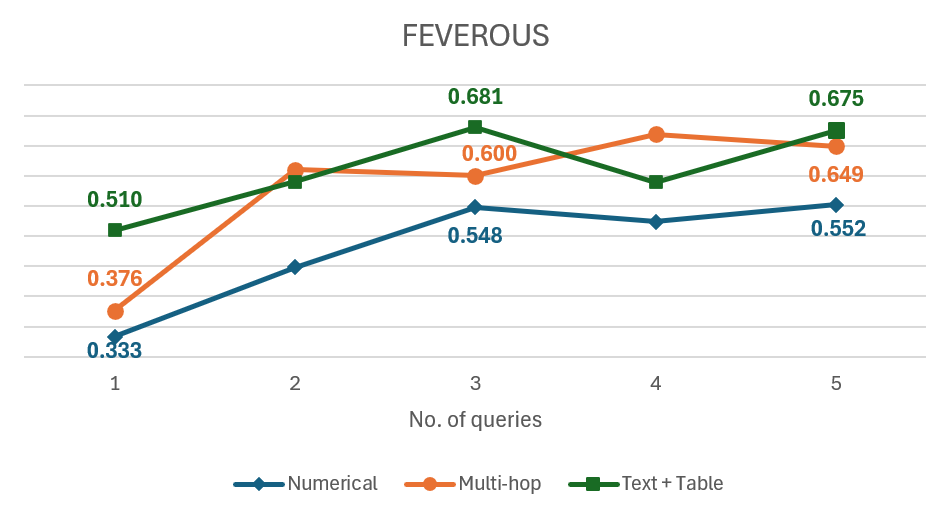}%
      \hfill
      \includegraphics[width=.33\linewidth]{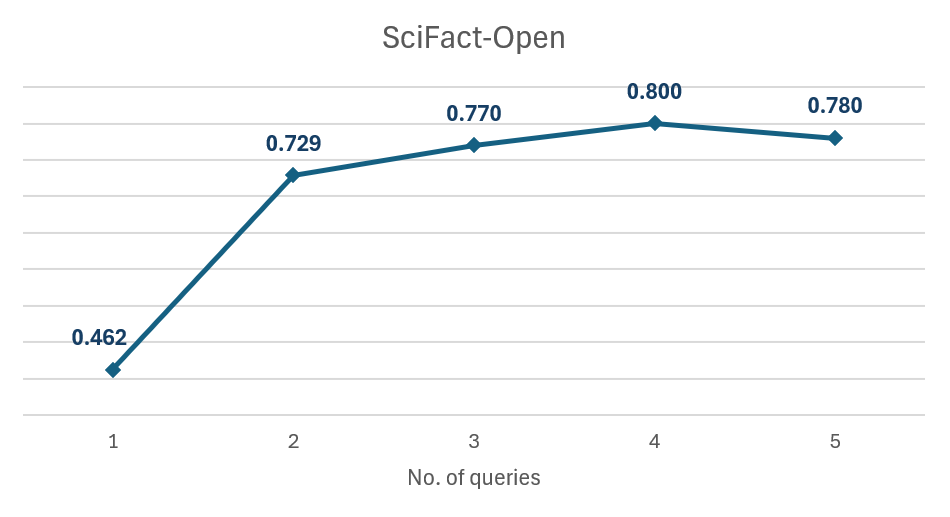}%
    \end{minipage}
  }
  \caption{Macro F1-score of varying the number of generated questions per claim from 1 to 5 for Query Generation Agent. The Macro F1-score generally increases as the number of generated questions per claim grows, peaking at 3 or 4 questions across most settings.}
  \label{tab:fine-tune-query}
\end{figure*}

To assess the influence of question generation granularity, Table~\ref{tab:fine-tune-query} presents results obtained by varying the number of sub-questions generated per claim from 1 to 5. The experiments were carried out using the GPT-4o-mini model, which showed the best overall performance in the previous evaluation.

The results show that increasing the number of generated questions enhances performance, particularly when scaling from 1 to 3 questions, with notable gains in the F1 score in HoVER 2-hop (0.472 to 0.600) and SciFact-Open (0.462 to 0.770). This suggests that a finer decomposition of claims improves the accuracy of evidence retrieval and verification. However, beyond three questions, the gains plateau or slightly decline, as seen in HoVER 2-hop (dropping from 0.600 at 3 questions to 0.538 at 5), likely due to redundant or noisy queries. For more complex tasks such as HoVER 4-hop and SciFact-Open, performance continues to improve up to 4 to 5 questions, indicating that deeper claims benefit from greater granularity. In general, generating 3–4 questions per claim provides the best balance between reasoning depth and evidence precision in multi-hop verification.
\subsection{Comparison with Open-source Models}
The results in Table~\ref{tab:results_llama} highlight the varying performance of the baseline methods across different models and tasks. For Llama-3.2-1B, CoT consistently outperforms all other methods in most tasks, demonstrating its ability to leverage intermediate reasoning effectively even with limited model capacity. However, more sophisticated methods such as FOLK and MAS struggle in this small architecture, producing scores close to randomness in many settings. This suggests that, for smaller models, adding complexity through multi-hop or multi-agent strategies may overwhelm their ability to perform robust reasoning. This phenomenon is consistent with findings in the literature, where smaller models exhibit degraded performance on complex reasoning tasks, such as question answering and natural language inference, due to limited computational capacity \cite{10.5555/3495724.3495883}.

With Qwen-2.5-3B, the results show greater variability and improvement for methods employing more sophisticated strategies. MAS performs strongly on 4-Hop and Numerical tasks, while SA+SE and FOLK outperform CoT in certain cases, reflecting the benefits of enhanced capacity for multi-hop and numerical reasoning. However, performance gains are not uniform across all tasks and methods, indicating that the utility of sophisticated strategies depends not only on the complexity of the task but also on the scale of the underlying model~\cite{10.5555/3495724.3495883}. Overall, these findings underscore the growing potential of multi-hop and multi-agent methods to outperform simpler baselines as model capacity increases.
\begin{table*}[ht]
\centering
\begin{adjustbox}{max width=\textwidth}
\begin{tabular}{l|ccc|ccc|c}
\toprule
\textbf{Method} & \multicolumn{3}{c|}{\textbf{HoVER}} & \multicolumn{3}{c|}{\textbf{FEVEROUS}} & \textbf{SciFact-Open} \\
& \textbf{2-Hop} & \textbf{3-Hop} & \textbf{4-Hop} & \textbf{Numerical} & \textbf{Multi-hop} & \textbf{Text+Table} &  \\
\midrule
\multicolumn{8}{c}{\textbf{Model: Llama-3.2-1B}} \\
\midrule
Direct   & 0.392 & 0.493 & 0.448 & 0.446 & 0.443 & 0.502 & \textbf{0.577} \\
CoT      & \textbf{0.519} & \textbf{0.499} & \textbf{0.510} & \textbf{0.463} & \textbf{0.513} & \textbf{0.540} & 0.559 \\
SA+SE & 0.315 & 0.369 & 0.329 & 0.333 & 0.333 & 0.333 & 0.367 \\
FOLK & 0.315 & 0.346 & 0.329 & 0.333 & 0.333 & 0.355 & 0.342 \\
\textbf{MAS (Ours)} & 0.315 & 0.369 & 0.329 & 0.329 & 0.333 & 0.333 & 0.365 \\
\midrule
\multicolumn{8}{c}{\textbf{Model: Qwen-2.5-3B}} \\
\midrule
Direct   & 0.387 & 0.350 & 0.338 & 0.355 & 0.423 & 0.416 & 0.625 \\
CoT      & 0.387 & 0.310 & 0.338 & 0.355 & 0.405 & 0.436 & 0.538 \\
SA+SE & \textbf{0.490} & 0.320 & 0.372 & 0.329 & \textbf{0.423} & 0.448 & \textbf{0.664} \\
FOLK & 0.448 & 0.369 & 0.402 & 0.500 & 0.366 & 0.345 & 0.508 \\
\textbf{MAS (Ours)} & 0.446 & \textbf{0.378 (+0.009)} & \textbf{0.493 (+0.091)} & \textbf{0.533 (+0.033)} & 0.363 & \textbf{0.502 (+0.054)} & 0.422 \\
\bottomrule
\end{tabular}
\end{adjustbox}
\caption{Macro F1-score of baseline methods and MAS across HoVER, FEVEROUS, and SciFact-Open datasets using Llama-3.2-1B and Qwen-2.5-3B. The table highlights the varying performance of methods as model capacity increases, with CoT outperforming on small models and sophisticated methods like MAS and SA+SE yielding strong results on Qwen-2.5-3B. The number of questions generated by the Query Generation Agent is 3.}
\label{tab:results_llama}
\end{table*}

\section{Analysis}
\subsection{The Impact of Filtering out Non-verifiable Statements}
\begin{figure}[htbp]
    \centering
    \includegraphics[width=\linewidth]{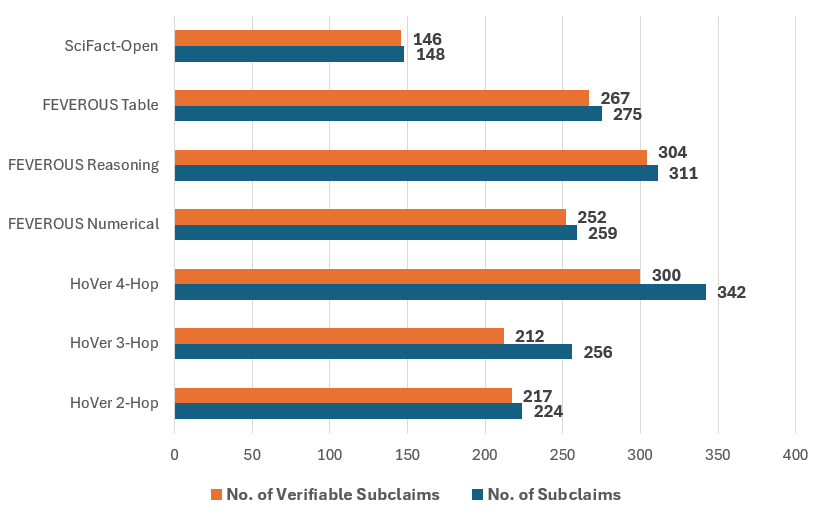}
    \caption{Number of subclaims and verifiable subclaims across different datasets. The number of subclaims increases with claim complexity, while the proportion of unverifiable subclaims remains low across most datasets.}
    \label{fig:subclaim-verifiable}
\end{figure}

Most subclaims generated across datasets are verifiable, indicating effective decomposition, although the rate of unverifiable components varies with claim complexity. Figure \ref{fig:subclaim-verifiable} shows that the number of subclaims grows with the complexity of the claim, reflecting the need for finer decomposition to handle multi-hop reasoning. The rate of unverifiable subclaims varies across datasets, with HoVer 3-Hop and 4-Hop having the highest proportions (up to 17.19\%) and FEVEROUS and SciFact-Open having much lower rates (under 3.13\%). This suggests that certain datasets naturally present more subjective or hard-to-verify components. Overall, the low proportion of unverifiable subclaims, even for complex cases, indicates that the decomposition process effectively breaks down most claims into components that can be objectively evaluated.
\subsection{The Impact of Source Credibility Assessment}
\begin{figure}[htbp]
    \centering
    \includegraphics[width=0.8\linewidth]{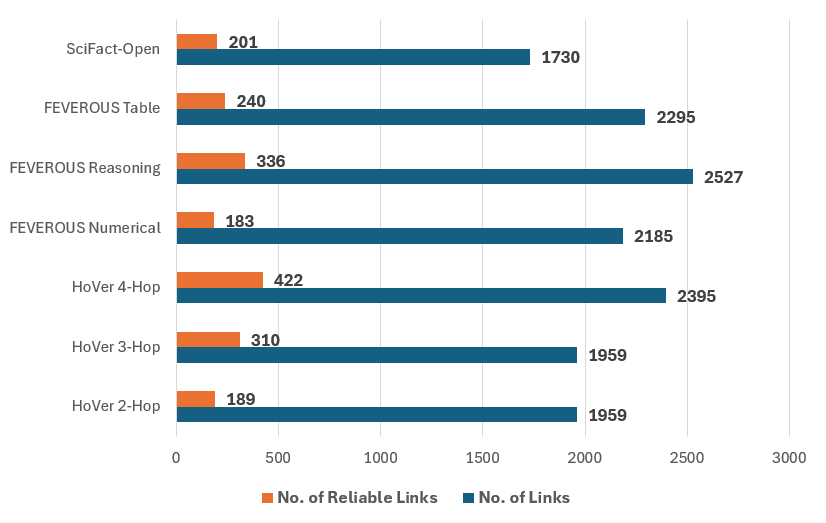}
    \caption{Comparison of total retrieved links and credible links after credibility filtering across datasets. The results highlight a significant reduction in usable links for multi-hop datasets, reflecting a higher prevalence of unreliable sources.}
    \label{fig:credit-check}
\end{figure}

Credibility filtering significantly reduces the number of usable links, especially for multi-hop datasets, reflecting a higher prevalence of unreliable sources in more complex tasks. Figure \ref{fig:credit-check} reveals that credibility filtering significantly reduces usable links across datasets, exposing the prevalence of low-credibility sources in open-domain retrieval. For HoVer, only 9.6\% of 2-hop and 15.8\% of 3-hop links pass filtering, emphasizing the need for robust credibility checks to prevent unreliable evidence propagation. Conversely, FEVEROUS and SciFact-Open retain over 86\% of links, indicating more trustworthy retrieval and stronger alignment with credible sources. These results highlight the critical role of credibility filtering in multi-hop fact verification and its varying impact across datasets and tasks.
\subsection{Assessing the Quality of Explanations}
MAS outperforms other methods by producing the most coherent and comprehensive explanations, topping 5 of 7 categories across HoVER, FEVEROUS, and SciFact-Open datasets. To evaluate the quality of explanations generated by FOLK, GPT-4 was chosen as the LLM-as-judge for ranking (1 for best, 4 for least preferred) due to its proven alignment with human judgment, outperforming other models in scoring tasks and surpassing human-to-human agreement consistency~\cite{10.5555/3666122.3668142}.  We instructed GPT-4 to compare and rank explanations produced by four different methods: CoT, Self-Ask, FOLK and MAS.
The evaluation focused on the following three criteria by methodology established by Atanasova et al. \cite{atanasova-etal-2020-generating-fact}:

\begin{itemize}
    \item \textbf{Coverage}: The extent to which the explanation includes all salient and relevant information necessary to verify the claim. Annotators were given access to gold-standard annotated evidence and asked to judge whether the explanation adequately captured the critical points present in this evidence.
    
    \item \textbf{Soundness}: The logical consistency of the explanation, ensuring that it does not contain contradictions with the claim or the provided gold evidence.
    
    \item \textbf{Readability}: The clarity and coherence of the explanation, measuring how easily a reader can understand the content. This criterion emphasizes the explanation’s linguistic quality and comprehensibility.
\end{itemize}

The explanation quality results in Table~\ref{tab:gpt-eval} show that MAS produces the most coherent and comprehensive explanations across most tasks. MAS tops 5 of 7 categories, particularly excelling in complex 2-Hop (1.73), 3-Hop (1.90), and 4-Hop (1.60) scenarios. FOLK performs well in specialized cases, securing the best ranks in FEVEROUS Multi-hop (1.95) and SciFact-Open (2.15), while CoT consistently ranks lowest (from 2.85 to 3.41). SA+SE shows moderate performance but falls short of MAS and FOLK. These results align with Table~\ref{tab:results}, confirming MAS's ability to produce more interpretable explanations alongside higher accuracy.
\begin{table*}[ht]
\centering
\begin{adjustbox}{width=0.9\textwidth}
\begin{tabular}{l|ccc|ccc|c}
\toprule
\textbf{Method} & \multicolumn{3}{c|}{\textbf{HoVER}} & \multicolumn{3}{c|}{\textbf{FEVEROUS}} & \textbf{SciFact-Open} \\
& \textbf{2-Hop} & \textbf{3-Hop} & \textbf{4-Hop} & \textbf{Numerical} & \textbf{Multi-hop} & \textbf{Text+Table} &  \\
\midrule
CoT & 3.41 & 3.15 & 3.28 & 3.01 & 2.85 & 2.95 & 2.87  \\
SA+SE &  2.64 & 2.75 & 2.82 & 2.49 & 2.65 & 2.85 & 2.73  \\
FOLK & 2.22 & 2.20 & 2.30 & 2.34 & \textbf{1.95} & 2.20 &\textbf{2.15} \\
\textbf{MAS (Ours)} & \textbf{1.73 (-0.49)} & \textbf{1.90 (-0.30)} & \textbf{1.60 (-0.70)} & \textbf{2.16 (-0.18)} & 2.55 & \textbf{2.00 (-0.20)} & 2.25 \\
\bottomrule
\end{tabular}
\end{adjustbox}
\caption{
Mean Average Ranks (MARs) of explanations, averaged across Coverage, Soundness, and Readability criteria, for HoVER, FEVEROUS, and SciFact-Open datasets. The lower MAR indicates a higher ranking and represents a better quality of an explanation. The best result per column is highlighted in \textbf{bold}. MAS consistently achieves the lowest MARs in most categories, excelling in complex multi-hop tasks, while FOLK leads in specialized datasets, and CoT ranks lowest across all tasks.}
\label{tab:gpt-eval}
\end{table*}

\section{Conclusion}
This paper introduces a novel multi-agent system for automated fact-checking, designed to enhance accuracy, efficiency, and explainability in combating online misinformation. The system features four specialized agents: Input Ingestion, Query Generation, Evidence Retrieval, and Verdict Prediction. This architecture enables structured reasoning and modular information processing, significantly improving the precision of fact-checking. A key contribution is the robust evidence retrieval strategy, which prioritizes source credibility and utilizes the full content of the document. Experimental evaluations on benchmark datasets like FEVEROUS, HOVER, and SciFact demonstrate the system's superior performance, achieving a 12.3\% relative improvement in Macro F1-score over baseline methods. 

In essence, this multi-agent system effectively addresses the limitations of existing automated fact-checking approaches by providing a more accurate, efficient, and transparent verification methodology. Its structured pipeline and emphasis on credible source retrieval offer a scalable solution for real-world fact-checking, aligning closely with human verification practices while providing interpretable outcomes.

\section{Limitations and Future Works}
Our multi-agent fact verification system currently faces three primary limitations. First, its reliance on the US-centric configuration of SerperAPI introduces geographic and cultural bias, limiting the effectiveness of claim verification in non-US. contexts or culturally nuanced scenarios.Second, the evaluation framework emphasizes quantitative metrics, but does not incorporate human evaluation to assess the qualitative aspects of explanations. Third, the overall architecture has its credibility filtering based on MBFC, which is constrained by limited coverage of international or domain-specific sources.

To address these challenges, future research should explore: (1) enhancing geographic and cultural representativeness through regional API routing, language-localized search, and multilingual query generation \cite{google2024international}; (2) incorporating human evaluations to assess explanation clarity and argument strength \cite{wang-shu-2023-explainable}; (3) designing more comprehensive credibility assessment frameworks that integrate trust signals, domain expertise, and dynamic source scoring \cite{HILLIGOSS20081467}; (4) extending the system to handle multimodal claims involving images, videos, or audio \cite{10.1007/978-3-030-58323-1_3}. These directions aim to build a more transparent and globally robust fact-checking system.

\bibliographystyle{IEEEtran}
\footnotesize\bibliography{ref}

\begin{thebibliography}{10}
\providecommand{\url}[1]{#1}
\csname url@samestyle\endcsname
\providecommand{\newblock}{\relax}
\providecommand{\bibinfo}[2]{#2}
\providecommand{\BIBentrySTDinterwordspacing}{\spaceskip=0pt\relax}
\providecommand{\BIBentryALTinterwordstretchfactor}{4}
\providecommand{\BIBentryALTinterwordspacing}{\spaceskip=\fontdimen2\font plus
\BIBentryALTinterwordstretchfactor\fontdimen3\font minus \fontdimen4\font\relax}
\providecommand{\BIBforeignlanguage}[2]{{%
\expandafter\ifx\csname l@#1\endcsname\relax
\typeout{** WARNING: IEEEtran.bst: No hyphenation pattern has been}%
\typeout{** loaded for the language `#1'. Using the pattern for}%
\typeout{** the default language instead.}%
\else
\language=\csname l@#1\endcsname
\fi
#2}}
\providecommand{\BIBdecl}{\relax}
\BIBdecl

\bibitem{10.1145/3442188.3445922}
A.~Bondielli and F.~Marcelloni, ``A survey on fake news and rumour detection techniques,'' \emph{Information Sciences}, vol. 497, pp. 38--55, 2020.

\bibitem{Graves2018UnderstandingTP}
L.~Graves, ``Understanding the promise and limits of automated fact-checking,'' \emph{Factsheet}, 2018.

\bibitem{doi:10.1126/science.aap9559}
D.~M.~J. Lazer, M.~A. Baum, Y.~Benkler, A.~J. Berinsky, K.~M. Greenhill, F.~Menczer, M.~J. Metzger, B.~Nyhan, G.~Pennycook, D.~Rothschild, M.~Schudson, S.~A. Sloman, C.~R. Sunstein, E.~A. Thorson, D.~J. Watts, and J.~L. Zittrain, ``The science of fake news,'' \emph{Science}, vol. 359, no. 6380, pp. 1094--1096, 2018.

\bibitem{doi:10.1177/0268580920914755}
M.~Cinelli, W.~Quattrociocchi, A.~Galeazzi, C.~M. Valensise, E.~Brugnoli, A.~L. Schmidt, P.~Zola, F.~Zollo, and A.~Scala, ``The covid-19 social media infodemic,'' \emph{Scientific Reports}, vol.~10, no.~1, p. 16598, 2020.

\bibitem{10.1016/j.ipm.2019.03.004}
K.~Shu, A.~Sliva, S.~Wang, J.~Tang, and H.~Liu, ``Fake news detection on social media: A data mining perspective,'' \emph{ACM SIGKDD Explorations Newsletter}, vol.~19, no.~1, pp. 22--36, 2017.

\bibitem{zhao2023survey}
W.~X. Zhao, K.~Zhou, J.~Li, T.~Tang, X.~Wang, Y.~Hou, Y.~Min, B.~Zhang, J.~Zhang, Z.~Dong \emph{et~al.}, ``A survey of large language models,'' \emph{arXiv preprint arXiv:2303.18223}, 2023.

\bibitem{10.1145/3571730}
Z.~Ji, N.~Lee, R.~Frieske, T.~Yu, D.~Su, Y.~Xu, E.~Ishii, Y.~Bang, A.~Madotto, and P.~Fung, ``Survey of hallucination in natural language generation,'' \emph{ACM Computing Surveys}, vol.~55, no.~12, pp. 1--38, 2023.

\bibitem{Feng2023KnowledgeST}
S.~Feng, W.~Jiang, C.~Graber, K.~Vafa, N.~Mathur, Y.~Zhou, B.~I.~P. Rubinstein, and J.~Leskovec, ``Knowledge staleness in large language models,'' \emph{arXiv preprint arXiv:2310.19215}, 2023.

\bibitem{lewis2020retrieval}
P.~Lewis, E.~Perez, A.~Piktus, F.~Petroni, V.~Karpukhin, N.~Goyal, H.~Küttler, M.~Lewis, W.-t. Yih, T.~Rocktäschel \emph{et~al.}, ``Retrieval-augmented generation for knowledge-intensive nlp tasks,'' in \emph{Advances in Neural Information Processing Systems}, vol.~33, 2020, pp. 9459--9474.

\bibitem{wei2022chain}
J.~Wei, X.~Wang, D.~Schuurmans, M.~Bosma, B.~Ichter, F.~Xia, E.~Chi, Q.~Le, and D.~Zhou, ``Chain of thought prompting elicits reasoning in large language models,'' \emph{arXiv preprint arXiv:2201.11903}, 2022.

\bibitem{chern2023factool}
B.~Chern, S.~Peng, A.~Verma, X.~Yin, A.~Ranade, N.~Jindal, R.~Joshi, A.~Saxena, and R.~Sarikaya, ``Factool: Factuality detection in generative ai,'' \emph{arXiv preprint arXiv:2307.13528}, 2023.

\bibitem{10.1145/3137597.3137600}
K.~Popat, S.~Mukherjee, J.~Strötgen, and G.~Weikum, ``Where the truth lies: Explaining the credibility of emerging claims on the web and social media,'' \emph{Proceedings of the 26th International Conference on World Wide Web Companion}, pp. 1003--1012, 2017.

\bibitem{Aly21Feverous}
\BIBentryALTinterwordspacing
R.~Aly, Z.~Guo, M.~S. Schlichtkrull, J.~Thorne, A.~Vlachos, C.~Christodoulopoulos, O.~Cocarascu, and A.~Mittal, ``{FEVEROUS}: Fact extraction and {VERification} over unstructured and structured information,'' in \emph{Thirty-fifth Conference on Neural Information Processing Systems Datasets and Benchmarks Track (Round 1)}, 2021. [Online]. Available: \url{https://openreview.net/forum?id=h-flVCIlstW}
\BIBentrySTDinterwordspacing

\bibitem{jiang-etal-2020-hover}
Y.~Jiang, S.~Bordia, Z.~Zhong, C.~Dognin, M.~Singh, and M.~Bansal, ``{HOVER}: A dataset for many-hop fact extraction and claim verification,'' in \emph{Findings of the Association for Computational Linguistics: EMNLP 2020}, 2020, pp. 3441--3460.

\bibitem{wadden-etal-2020-fact}
D.~Wadden, S.~Lin, K.~Lo, L.~L. Wang, M.~van Zuylen, A.~Cohan, and H.~Hajishirzi, ``Fact or fiction: Verifying scientific claims,'' in \emph{Proceedings of the 2020 Conference on Empirical Methods in Natural Language Processing (EMNLP)}, 2020, pp. 7534--7550.

\bibitem{politifact}
PolitiFact, ``Politifact,'' \url{https://www.politifact.com/}, accessed: 2025-03-20.

\bibitem{snopes}
Snopes, ``Snopes,'' \url{https://www.snopes.com/}, accessed: 2025-03-20.

\bibitem{factcheck}
FactCheck, ``Factcheck,'' \url{https://www.factcheck.org/}, accessed: 2025-03-20.

\bibitem{amazeen2015revisiting}
M.~A. Amazeen, ``Revisiting the epistemology of fact-checking,'' \emph{Critical Review}, vol.~27, no.~1, pp. 1--22, 2015.

\bibitem{Vlachos2014FactCT}
A.~Vlachos and S.~Riedel, ``Fact checking: Task definition and dataset construction,'' in \emph{Proceedings of the ACL 2014 Workshop on Language Technologies and Computational Social Science}, 2014, pp. 18--22.

\bibitem{hassan2017toward}
N.~Hassan, G.~Zhang, F.~Arslan, J.~Caraballo, D.~Jimenez, S.~Gawsane, S.~Hasan, M.~Joseph, A.~Kulkarni, A.~K. Nayak \emph{et~al.}, ``Toward automated fact-checking: Detecting check-worthy factual claims by claimbuster,'' in \emph{Proceedings of the 23rd ACM SIGKDD International Conference on Knowledge Discovery and Data Mining}, 2017, pp. 1803--1812.

\bibitem{popat2018declare}
K.~Popat, S.~Mukherjee, A.~Yates, and G.~Weikum, ``Declare: Debunking fake news and false claims using evidence-aware deep learning,'' in \emph{Proceedings of the 2018 Conference on Empirical Methods in Natural Language Processing}, 2018, pp. 22--32.

\bibitem{augenstein-etal-2019-multifc}
I.~Augenstein, C.~Lioma, D.~Wang, L.~C. Lima, C.~Hansen, C.~Hansen, and J.~G. Pedersen, ``{MultiFC}: A real-world multi-domain dataset for evidence-based fact checking of claims,'' in \emph{Proceedings of the 2019 Conference on Empirical Methods in Natural Language Processing and the 9th International Joint Conference on Natural Language Processing (EMNLP-IJCNLP)}, 2019, pp. 4685--4697.

\bibitem{zhou-etal-2019-gear}
J.~Zhou, X.~Han, C.~Yang, Z.~Liu, L.~Wang, C.~Li, and M.~Sun, ``{GEAR}: Graph-based evidence aggregating and reasoning for fact verification,'' in \emph{Proceedings of the 57th Annual Meeting of the Association for Computational Linguistics}, 2019, pp. 892--901.

\bibitem{Stiff2021DetectingCD}
C.~Stiff and F.~Johansson, ``Detecting covid-19 misinformation with veracity assessment of social media posts via deep learning,'' \emph{Journal of Medical Internet Research}, vol.~23, no.~9, p. e30315, 2021.

\bibitem{lee-etal-2021-towards}
N.~Lee, B.~Z. Li, S.~Wang, W.-t. Yih, H.~Ma, and M.~Khabsa, ``Towards few-shot fact-checking via perplexity,'' in \emph{Proceedings of the 2021 Conference of the North American Chapter of the Association for Computational Linguistics: Human Language Technologies}, 2021, pp. 1971--1981.

\bibitem{DU2023QuantifyingAA}
Y.~Du, S.~Ding, Z.~Zhao, Y.~Lin, R.~Nallapati, B.~Xiang, B.~Zhou, D.~Roth, L.~Zettlemoyer, X.~Liang \emph{et~al.}, ``Quantifying and analyzing hallucinations in large language models,'' \emph{arXiv preprint arXiv:2310.00905}, 2023.

\bibitem{xi2023rise}
Z.~Xi, W.~Chen, X.~Guo, W.~He, Y.~Ding, X.~Hong, Y.~Luo, W.~Liang, L.~Bing, L.~Si \emph{et~al.}, ``The rise and potential of large language model based agents: A survey,'' \emph{arXiv preprint arXiv:2309.07864}, 2023.

\bibitem{li2023can}
Y.~Li and K.~Shu, ``Can large language models provide faithful explanations for fake news detection?'' \emph{arXiv preprint arXiv:2305.15005}, 2023.

\bibitem{zhang2023web}
X.~Zhang, H.~Jiang, W.~Yin, X.~Ren, and J.~Han, ``Web agents: Evaluating the reliability of web search engines for ai agents,'' \emph{arXiv preprint arXiv:2312.09254}, 2023.

\bibitem{wang2023self}
M.~Wang, Z.~Yin, M.~Guo, X.~Jiang, X.~Ren, and J.~Han, ``Self-checker: Plug-and-play modules for fact-checking with large language models,'' \emph{arXiv preprint arXiv:2305.14623}, 2023.

\bibitem{enderton2001mathematical}
H.~B. Enderton, \emph{A mathematical introduction to logic}.\hskip 1em plus 0.5em minus 0.4em\relax Elsevier, 2001.

\bibitem{wang-shu-2023-explainable}
\BIBentryALTinterwordspacing
H.~Wang and K.~Shu, ``Explainable claim verification via knowledge-grounded reasoning with large language models,'' in \emph{Findings of the Association for Computational Linguistics: EMNLP 2023}, H.~Bouamor, J.~Pino, and K.~Bali, Eds.\hskip 1em plus 0.5em minus 0.4em\relax Singapore: Association for Computational Linguistics, Dec. 2023, pp. 6288--6304. [Online]. Available: \url{https://aclanthology.org/2023.findings-emnlp.416/}
\BIBentrySTDinterwordspacing

\bibitem{micallef2022claim}
D.~Micallef, K.~Kakaes, A.~Haghighi, S.~Lightseed, L.~Gu, Q.~Liao, I.~Liskovich, Y.~W. Tay, and E.~Kamar, ``Claim matching beyond english to scale global fact-checking,'' in \emph{Proceedings of the 60th Annual Meeting of the Association for Computational Linguistics (Volume 1: Long Papers)}.\hskip 1em plus 0.5em minus 0.4em\relax Association for Computational Linguistics, 2022, pp. 5249--5264.

\bibitem{konstantinovskiy2021claim}
L.~Konstantinovskiy, O.~Price, M.~Babakar, and A.~Zubiaga, ``Claim detection in biomedical twitter posts,'' in \emph{Proceedings of the 20th Workshop on Biomedical Language Processing}.\hskip 1em plus 0.5em minus 0.4em\relax Association for Computational Linguistics, 2021, pp. 131--142.

\bibitem{press-etal-2023-measuring}
\BIBentryALTinterwordspacing
O.~Press, M.~Zhang, S.~Min, L.~Schmidt, N.~Smith, and M.~Lewis, ``Measuring and narrowing the compositionality gap in language models,'' in \emph{Findings of the Association for Computational Linguistics: EMNLP 2023}, H.~Bouamor, J.~Pino, and K.~Bali, Eds.\hskip 1em plus 0.5em minus 0.4em\relax Singapore: Association for Computational Linguistics, Dec. 2023, pp. 5687--5711. [Online]. Available: \url{https://aclanthology.org/2023.findings-emnlp.378/}
\BIBentrySTDinterwordspacing

\bibitem{baye2016search}
M.~R. Baye, B.~De~los Santos, and M.~R. Wildenbeest, ``Search engine optimization: what drives organic traffic to retail sites?'' \emph{Journal of Economics \& Management Strategy}, vol.~25, no.~1, pp. 6--31, 2016.

\bibitem{7977137}
S.~Krrabaj, F.~Baxhaku, and D.~Sadrijaj, ``Investigating search engine optimization techniques for effective ranking: A case study of an educational site,'' in \emph{2017 6th Mediterranean Conference on Embedded Computing (MECO)}, 2017, pp. 1--4.

\bibitem{serper2024}
{Serper}, ``Serper: Real-time google search api,'' \url{https://serper.dev}, 2024, accessed: 2025-04-05.

\bibitem{mediabiasfactcheck}
``Media bias/fact check api,'' \url{https://mediabiasfactcheck.com/}, accessed: 2025-04-18.

\bibitem{METZGER2013210}
\BIBentryALTinterwordspacing
M.~J. Metzger and A.~J. Flanagin, ``Credibility and trust of information in online environments: The use of cognitive heuristics,'' \emph{Journal of Pragmatics}, vol.~59, pp. 210--220, 2013, biases and constraints in communication: Argumentation, persuasion and manipulation. [Online]. Available: \url{https://www.sciencedirect.com/science/article/pii/S0378216613001768}
\BIBentrySTDinterwordspacing

\bibitem{dong2015knowledge}
X.~L. Dong, E.~Gabrilovich, K.~Murphy, V.~Dang, W.~Horn, C.~Lugaresi, S.~Sun, and W.~Zhang, ``Knowledge-based trust: Estimating the trustworthiness of web sources,'' in \emph{Proceedings of the VLDB Endowment}, vol.~8, no.~9.\hskip 1em plus 0.5em minus 0.4em\relax VLDB Endowment, 2015, pp. 938--949.

\bibitem{Zhou2024CorrectingMO}
\BIBentryALTinterwordspacing
X.~Zhou, A.~Sharma, A.~X. Zhang, and T.~Althoff, ``Correcting misinformation on social media with a large language model,'' \emph{ArXiv}, vol. abs/2403.11169, 2024. [Online]. Available: \url{https://api.semanticscholar.org/CorpusID:268513555}
\BIBentrySTDinterwordspacing

\bibitem{zhang-gao-2023-towards}
\BIBentryALTinterwordspacing
X.~Zhang and W.~Gao, ``Towards {LLM}-based fact verification on news claims with a hierarchical step-by-step prompting method,'' in \emph{Proceedings of the 13th International Joint Conference on Natural Language Processing and the 3rd Conference of the Asia-Pacific Chapter of the Association for Computational Linguistics (Volume 1: Long Papers)}, J.~C. Park, Y.~Arase, B.~Hu, W.~Lu, D.~Wijaya, A.~Purwarianti, and A.~A. Krisnadhi, Eds.\hskip 1em plus 0.5em minus 0.4em\relax Nusa Dua, Bali: Association for Computational Linguistics, Nov. 2023, pp. 996--1011. [Online]. Available: \url{https://aclanthology.org/2023.ijcnlp-main.64/}
\BIBentrySTDinterwordspacing

\bibitem{selenium2023}
{Selenium Contributors}, ``Selenium: Web browser automation,'' \url{https://www.selenium.dev/}, 2023, accessed: 2025-04-24.

\bibitem{richardson2007beautiful}
L.~Richardson, ``Beautiful soup documentation,'' 2007.

\bibitem{team2024gemini}
G.~Team, P.~Georgiev, V.~I. Lei, R.~Burnell, L.~Bai, A.~Gulati, G.~Tanzer, D.~Vincent, Z.~Pan, S.~Wang \emph{et~al.}, ``Gemini 1.5: Unlocking multimodal understanding across millions of tokens of context,'' \emph{arXiv preprint arXiv:2403.05530}, 2024.

\bibitem{10.5555/3600270.3602070}
J.~Wei, X.~Wang, D.~Schuurmans, M.~Bosma, B.~Ichter, F.~Xia, E.~H. Chi, Q.~V. Le, and D.~Zhou, ``Chain-of-thought prompting elicits reasoning in large language models,'' in \emph{Proceedings of the 36th International Conference on Neural Information Processing Systems}, ser. NIPS '22.\hskip 1em plus 0.5em minus 0.4em\relax Red Hook, NY, USA: Curran Associates Inc., 2022.

\bibitem{10.5555/3495724.3495883}
T.~B. Brown, B.~Mann, N.~Ryder, M.~Subbiah, J.~Kaplan, P.~Dhariwal, A.~Neelakantan, P.~Shyam, G.~Sastry, A.~Askell, S.~Agarwal, A.~Herbert-Voss, G.~Krueger, T.~Henighan, R.~Child, A.~Ramesh, D.~M. Ziegler, J.~Wu, C.~Winter, C.~Hesse, M.~Chen, E.~Sigler, M.~Litwin, S.~Gray, B.~Chess, J.~Clark, C.~Berner, S.~McCandlish, A.~Radford, I.~Sutskever, and D.~Amodei, ``Language models are few-shot learners,'' in \emph{Proceedings of the 34th International Conference on Neural Information Processing Systems}, ser. NIPS '20.\hskip 1em plus 0.5em minus 0.4em\relax Red Hook, NY, USA: Curran Associates Inc., 2020.

\bibitem{10.5555/3666122.3668142}
L.~Zheng, W.-L. Chiang, Y.~Sheng, S.~Zhuang, Z.~Wu, Y.~Zhuang, Z.~Lin, Z.~Li, D.~Li, E.~P. Xing, H.~Zhang, J.~E. Gonzalez, and I.~Stoica, ``Judging llm-as-a-judge with mt-bench and chatbot arena,'' in \emph{Proceedings of the 37th International Conference on Neural Information Processing Systems}, ser. NIPS '23.\hskip 1em plus 0.5em minus 0.4em\relax Red Hook, NY, USA: Curran Associates Inc., 2023.

\bibitem{atanasova-etal-2020-generating-fact}
\BIBentryALTinterwordspacing
P.~Atanasova, J.~G. Simonsen, C.~Lioma, and I.~Augenstein, ``Generating fact checking explanations,'' in \emph{Proceedings of the 58th Annual Meeting of the Association for Computational Linguistics}, D.~Jurafsky, J.~Chai, N.~Schluter, and J.~Tetreault, Eds.\hskip 1em plus 0.5em minus 0.4em\relax Online: Association for Computational Linguistics, Jul. 2020, pp. 7352--7364. [Online]. Available: \url{https://aclanthology.org/2020.acl-main.656/}
\BIBentrySTDinterwordspacing

\bibitem{google2024international}
\BIBentryALTinterwordspacing
Google, ``Managing multi-regional and multilingual sites,'' 2024, accessed 15 June 2025. [Online]. Available: \url{https://developers.google.com/search/docs/specialty/international/managing-multi-regional-sites}
\BIBentrySTDinterwordspacing

\bibitem{HILLIGOSS20081467}
\BIBentryALTinterwordspacing
B.~Hilligoss and S.~Y. Rieh, ``Developing a unifying framework of credibility assessment: Construct, heuristics, and interaction in context,'' \emph{Information Processing \& Management}, vol.~44, no.~4, pp. 1467--1484, 2008. [Online]. Available: \url{https://www.sciencedirect.com/science/article/pii/S0306457307002038}
\BIBentrySTDinterwordspacing

\bibitem{10.1007/978-3-030-58323-1_3}
\BIBentryALTinterwordspacing
A.~Giachanou, G.~Zhang, and P.~Rosso, ``Multimodal fake news detection with textual, visual and semantic information,'' in \emph{Text, Speech, and Dialogue: 23rd International Conference, TSD 2020, Brno, Czech Republic, September 8–11, 2020, Proceedings}.\hskip 1em plus 0.5em minus 0.4em\relax Berlin, Heidelberg: Springer-Verlag, 2020, p. 30–38. [Online]. Available: \url{https://doi.org/10.1007/978-3-030-58323-1_3}
\BIBentrySTDinterwordspacing

\end{thebibliography}

{\appendices
\clearpage
\onecolumn
\section*{Prompt for MAS}
\subsection*{Claim Decomposition Prompt}
The claim decomposition prompt instructs the language model to break down complex claims into atomic predicates using FOL representation. This prompt is used by the Input Ingestion Agent to transform natural language claims into verifiable components.

\lstset{
  style=cleantext,
  caption={Claim Decomposition Prompt. This prompt breaks complex claims into atomic predicates using FOL representation, enabling systematic verification by isolating individual factual assertions. The structured format facilitates comprehensive fact-checking through clear predicate identification.},
  label={fig:decomposition_prompt}
}

\begin{lstlisting}
You are given a problem description and a claim. The task is to define all the predicates in the claim and return them in JSON format, as shown in the example below.
Below is the example Claim: Howard University Hospital and Providence Hospital are both located in Washington, D.C.
{ "response": "Predicates:
Location(Howard_University_Hospital, Washington_D.C.) ::: Verify Howard University Hospital is located in Washington, D.C. 
Location(Providence_Hospital, Washington_D.C.) ::: Verify Providence Hospital is located in Washington, D.C. 

Below is the Claim: In 1959, former Chilean boxer Alfredo Cornejo Cuevas (born June 6, 1933) won the gold medal in the welterweight division at the Pan American Games (held in Chicago, United States, from August 27 to September 7) in Chicago, United States, and the world amateur welterweight title in Mexico City.
{ "response": "Predicates:
Born(Alfredo_Cornejo_Cuevas, June 6 1933) ::: Verify that Alfredo Cornejo Cuevas was born June 6, 1933. 
Won(Alfredo_Cornejo_Cuevas, the gold medal in the welterweight division at the Pan American Games in 1959) ::: Verify that Alfredo Cornejo Cuevas won the gold medal in the welterweight division at the Pan American Games in 1959. 
Held(The Pan American Games in 1959, Chicago United States) ::: Verify that the Pan American Games in 1959 were held in Chicago, United States. 
Won(Alfredo_Cornejo_Cuevas, the world amateur welterweight title in Mexico City) ::: Verify that Alfredo Cornejo Cuevas won the world amateur welterweight title in Mexico City. 
\end{lstlisting}
\subsection*{Subclaim Classification Prompt}
The subclaim classification prompt guides the language model in distinguishing between verifiable and non-verifiable claims. This prompt is utilized by the Input Ingestion Agent to filter out claims that cannot be objectively verified.

\lstset{
  style=cleantext,
  caption={Subclaim Classification Prompt. This prompt distinguishes between verifiable and non-verifiable claims using explicit criteria, filtering out subjective opinions, vague statements, and future predictions. This step ensures fact-checking resources are directed only toward claims that can be objectively verified.},
  label{fig:verifiability_prompt}
}

\begin{lstlisting}
You are an expert in claim verification. Your task is to determine whether a given claim is verifiable or non-verifiable.
A verifiable claim is a factual statement that can be checked against objective evidence from reliable sources. It makes specific assertions about the world that can be proven true or false through investigation.

A non-verifiable claim is one that cannot be objectively verified because it:
- Expresses a subjective opinion, preference, or personal experience  
- Makes vague or ambiguous statements without specific details  
- Refers to future events that haven't occurred yet  
- Makes normative or ethical judgments about what "should" be  
- Contains hypothetical scenarios or counterfactuals  
Examples:  
Verifiable: "The average global temperature increased by 0.8$^\circ$C between 1880 and 2012." 
Non-verifiable: "Climate change is the most important issue facing humanity today."  
Verifiable: "The film 'Parasite' won the Academy Award for Best Picture in 2020."  
Non-verifiable: "Parasite deserved to win the Academy Award for Best Picture."  
Please analyze the following claim and classify it as either VERIFIABLE or NON-VERIFIABLE. Provide a brief explanation for your classification.
Claim: {claim}  
Classification:
\end{lstlisting}


\subsection*{Query Generation Prompt}
The query generation prompt directs the Query Generation Agent in transforming atomic subclaims into effective search queries. This prompt is designed to maximize the retrieval of relevant evidence from search engines.

\lstset{
  style=cleantext,
  caption={Query Generation Prompt. This prompt transforms subclaims into diverse search queries using specific keywords, synonyms, and varying levels of specificity. This approach ensures comprehensive evidence gathering by exploring multiple angles while maintaining query relevance to the verification task.},
  label{fig:query_gen_prompt}
}

\begin{lstlisting}
For each input subclaim, generate k Google search question(s) that could be used to find evidence to verify the subclaim.
The questions should be diverse, exploring different aspects or perspectives related to the subclaim, while remaining clear and concise. Follow these guidelines:
1. Use Specific Keywords: Include precise terms related to entities and relationships in the claim.
2. Incorporate Synonyms and Related Terms: Use alternative phrasings to overcome vocabulary mismatches.
3. Vary Specificity: Generate both specific queries targeting exact details and broader queries that may capture contextual information.
4. Consider Different Angles: Approach the claim from multiple perspectives to ensure comprehensive evidence gathering.
5. Maintain Simplicity: Keep questions straightforward and directly relevant to the claim.

Return the output in JSON format like this: 
[{ 
    "claim": "Location(Howard Hospital, Washington D.C.) ::: Verify Howard University Hospital is located in Washington, D.C.", 
    "questions": ["Where is Howard Hospital located?"] 
}]
}
\end{lstlisting}


\subsection*{Content Retrieval Prompt}
The content retrieval prompt is used by the Evidence Seeking Agent to extract relevant information from web content. This prompt helps to filter lengthy web pages to identify only passages that directly address the verification query.
\lstset{
  style=cleantext,
  caption={Content Retrieval Prompt. This prompt extracts only information directly relevant to verification queries from potentially lengthy web content. This focused approach improves efficiency by filtering out extraneous information and ensuring that only pertinent evidence is considered in the verification process.},
  label{fig:retrieval_prompt}
}

\begin{lstlisting}
You are a helpful assistant who extracts information from text.
Given the following query and text content, extract only the sentences or phrases that directly
relate to the query. Do not include any information that is not relevant.
If the content contains no relevant information, return None.

Query: {query}

Content:
{content}

Relevant Information:
\end{lstlisting}




\subsection*{Verdict Prediction Prompt}
The verdict prediction prompt guides the Veracity Prediction Agent in synthesizing evidence and determining whether a subclaim is supported by the retrieved information. This prompt represents the final stage of the fact-checking pipeline.

\lstset{
  style=cleantext,
  caption={Verdict Prediction Prompt. This prompt synthesizes the evidence collected to determine the support for the subclaim through systematic analysis of credibility, consistency, and reliability. The structured decision-making process ensures justified verdicts with clear explanations based on the weight of evidence.},
  label{fig:verdict_prompt}
}

\begin{lstlisting}
You are an AI assistant responsible for determining whether a subclaim is supported by retrieved evidence.  
 
    ## Provided Information:
    This is a claim to do fact-checking:  
    \\n {claim}
    Here is the given subclaims, its subquestions, and retrieved evidence for each subquestion:  
    \\n {cell}  

    ## Decision-Making Process:

    1. Analyze the Retrieved Evidence  
    - Review all provided evidence relevant to the subclaim.  
    - Assess the credibility, consistency, and reliability of each piece of evidence.  

    2. Apply a Voting System for Classification  
    - If multiple sources strongly support the subclaim, classify it as "supported".  
    - If multiple sources contradict the subclaim, classify it as "not_supported".  
    - If the evidence is mixed, insufficient, or inconclusive, classify it as "not_supported".  

    3. Provide a Justification  
    - Clearly explain why the subclaim is classified as "supported" or "not_supported".  
    - Reference key pieces of evidence that influenced your decision.  
    - If the evidence is inconclusive, explain the limitations or uncertainties.  
    - Remember to adjust not to include " for later parse
    ## Response Format:
    Your response must be a structured JSON object:  

    ```json
    {{
        "label": "supported" or "not_supported",
        "explanation": "A concise, evidence-based summary supporting your decision."
    }}
\end{lstlisting}

\section*{Prompt for Explanation Evaluation}
The prompt is used by GPT-4 for the explanation evaluation task.

\lstset{
  style=cleantext,
  caption={Explanation Evaluation Prompt. This prompt assesses fact-checking explanations across multiple methods using coverage, soundness, and readability criteria. This comparative ranking framework provides quantitative measures of explanation quality while acknowledging potential differences in verdict determination.},
  label={fig:explain-eval}
}

\begin{lstlisting}
You are an expert evaluator for automated fact-check explanations. Your task is to:

- Review the original claim, its label, and the explanations produced by 4 methods. Each method may produce a different label; consider this when evaluating Soundness.  
- Evaluate each explanation according to 3 criteria:

  1. Coverage: To what extent the explanation includes all the salient and relevant information necessary to verify the claim.  
  2. Soundness: The logical consistency of the explanation; whether it supports or contradicts its own label and the original claim.  
  3. Readability: The clarity and coherence of the explanation; how easily a human can follow and understand it.

- Provide a **ranking (1 for best, 4 for worst)** for each criterion.  
Here is the input:
{
  "original_claim": "",
  "explanations": {
    "CoT": { "label": "<label>", "explanation": "<explanation>" }, 
    "Self-Ask": { "label": "<label>", "explanation": "<explanation>" },
    "FOLK": { "label": "<label>", "explanation": "<explanation>" }, 
    "MAS": { "label": "<label>", "explanation": "<explanation>" }
  }
}
The output should be in the format 
{
  "ranking": {
    "Coverage": { "1": "<method>", "2": "<method>", "3": "<method>", "4": "<method>" },
    "Soundness": { "1": "<method>", "2": "<method>", "3": "<method>", "4": "<method>" },
    "Readability": { "1": "<method>", "2": "<method>", "3": "<method>", "4": "<method>" }
  }
}
\end{lstlisting}

}

\end{document}